\documentclass[conference]{IEEEtran}
\IEEEoverridecommandlockouts
\usepackage{cite}
\usepackage{amsmath,amssymb,amsfonts}
\usepackage{algorithmic}
\usepackage{graphicx}
\usepackage{textcomp}
\usepackage{xcolor}
\usepackage{listings}
\usepackage{todonotes}
\usepackage{comment}
\usepackage{url}
\usepackage{float}
\usepackage{booktabs}
\usepackage{fancyvrb}
\usepackage{tablefootnote}

\usepackage{todonotes} % for comments

\def\BibTeX{{\rm B\kern-.05em{\sc i\kern-.025em b}\kern-.08em
    T\kern-.1667em\lower.7ex\hbox{E}\kern-.125emX}}
\begin{document}

\title{Spider4SPARQL: A Complex Benchmark for Evaluating Knowledge Graph Question Answering Systems
}

\author{\IEEEauthorblockN{Catherine Kosten}
\IEEEauthorblockA{%\textit{School of Engineering} \\
\textit{Zurich University of Applied Sciences}\\
Winterthur, Switzerland \\
Catherine.Kosten@zhaw.ch}
\and
\IEEEauthorblockN{ Philippe Cudré-Mauroux}
\IEEEauthorblockA{%\textit{Department of Informatics} \\
\textit{University of Fribourg}\\
Fribourg, Switzerland \\
Philippe.Cudre-Mauroux@unifr.ch}
\and
\IEEEauthorblockN{Kurt Stockinger}
\IEEEauthorblockA{%\textit{School of Engineering} \\
\textit{Zurich University of Applied Sciences}\\
Winterthur, Switzerland \\
Kurt.Stockinger@zhaw.ch}}
\begin{comment}
\and
\IEEEauthorblockN{4\textsuperscript{th} Given Name Surname}
\IEEEauthorblockA{\textit{dept. name of organization (of Aff.)} \\
\textit{name of organization (of Aff.)}\\
City, Country \\
email address or ORCID}
\and
\IEEEauthorblockN{5\textsuperscript{th} Given Name Surname}
\IEEEauthorblockA{\textit{dept. name of organization (of Aff.)} \\
\textit{name of organization (of Aff.)}\\
City, Country \\
email address or ORCID}
\and
\IEEEauthorblockN{6\textsuperscript{th} Given Name Surname}
\IEEEauthorblockA{\textit{dept. name of organization (of Aff.)} \\
\textit{name of organization (of Aff.)}\\
City, Country \\
email address or ORCID}
\end{comment}

\maketitle

\begin{abstract}
With the recent spike in the number and availability of Large Language Models (LLMs), it has become increasingly important to provide large and realistic benchmarks for evaluating Knowledge Graph Question Answering (KGQA) systems. So far the majority of benchmarks rely on pattern-based SPARQL query generation approaches. The subsequent natural language (NL) question generation is conducted through crowdsourcing or other automated methods, such as rule-based paraphrasing or NL question templates. Although some of these datasets are of considerable size, their pitfall lies in their pattern-based generation approaches, which do not always generalize well to the vague and linguistically diverse questions asked by humans in real-world contexts. In this paper, we introduce Spider4SPARQL - a new SPARQL benchmark dataset featuring 9,693 previously existing manually generated NL questions and 4,721 unique, novel, and complex SPARQL queries of varying complexity. In addition to the NL/SPARQL pairs, we also provide their corresponding 166 knowledge graphs and ontologies, which cover 138 different domains. Our complex benchmark enables novel ways of evaluating the strengths and weaknesses of modern KGQA systems. We evaluate the system with state-of-the-art KGQA systems as well as LLMs, which achieve only up to 45\% execution accuracy, demonstrating that Spider4SPARQL is a challenging benchmark for future research. 
\end{abstract}

\begin{IEEEkeywords}
Benchmark for Question Answering over Knowledge Graphs, Language Models, Performance Evaluation
\end{IEEEkeywords}
\section{Introduction}
\label{sec:intro}
Building systems for querying databases or knowledge graphs in natural language has been an important research topic over the last few decades \cite{copestake_jones_1990,EarlyWork2,EarlyWork3, affolter2019comparative}. Typical examples of such Text-to-SQL (also known as NL-to-SQL) or Text-to-SPARQL (also known as Knowledge Graph Question Answering (KGQA)) systems either use rule-based or machine learning-based approaches \cite{sima2021bio,brunner2021valuenet, katsogiannis2021deep}. The recent success of large language models has even further accelerated the race of building such systems as well as the need for datasets specifically designed for tasks like natural language (NL) to query language translation. One of the first benchmarks designed for translating natural language to a query language was WikiSQL \cite{WikiSQL}. Soon after, benchmarks for other query languages emerged, such as LC-QuAD 1.0 \cite{lcquad1.0} comprised of 5,000 NL/SPARQL pairs. 

Although these benchmarks were a step in the right direction, they lack queries that match the complexity and technical difficulty of queries from today's real-world knowledge graph applications. For example, LC-QuAD 1.0 queries only support single projections in the SELECT statement and COUNT aggregations, while WikiSQL only supports queries on a single table. 

LC-QuAD 2.0\cite{lcquad2.0} was the answer to the simplicity and small-scale of LC-QuAD 1.0. Although it is significantly larger than the original LC-QuAD dataset containing 30,000 NL/SPARQL pairs, it still lacks the necessary complexity that would allow this dataset to be used to train a real-world natural language interface for knowledge graphs.  Moreover, an evaluation of DBNQA \cite{DBLP:journals/corr/abs-1906-09302}, which has 894,499 NL/SPARQL pairs, shows that despite its impressive size, it still lacks both NL question and SPARQL query complexity and has a comparatively small vocabulary size. Real-world applications need to be able to execute queries that use different types of aggregations, different set operations and most importantly, multi-hop queries. 

As of today, there is a clear lack of KGQA benchmark datasets that generalize well to many domains and contain queries with the necessary complexity for training systems that can be used in real-world situations. Hence, we introduce Spider4SPARQL - a \textit{new SPARQL benchmark dataset featuring 10,181 previously existing manually generated NL questions and 5,693 unique, novel, and complex SPARQL queries of varying complexity}. Our benchmark is based on Spider \cite{Spider}, which has become the de-facto benchmark for evaluating Text-to-SQL systems and has accumulated over 100 submissions to its leaderboard\footnote{See Spider 1.0: Yale Semantic Parsing and Text-to-SQL Challenge \url{https://yale-lily.github.io/spider}}. The Spider dataset comprises 10,181 NL/SQL pairs created from 200 distinct databases and covers a range of topics from flights to restaurants. 

Our novel benchmark dataset Spider4SPARQL is built on the publicly available databases (train set, dev set) and 9,693 NL/SQL pairs from the Spider dataset. We have automatically translated Spider's NL/SQL pairs into NL/SPARQL pairs that can be executed on our 166 automatically generated knowledge graphs. The generation of ontologies and SPARQL queries uses an ontology-based data access (OBDA) approach\cite{PLCD*08} that mitigates the data modeling issues of the original Spider dataset. We make the dataset available on Github\footnote{\url{https://github.com/ckosten/Spider4SPARQL}}.
Our contributions are as follows:
\begin{itemize}
    \item We perform an \emph{analysis of the current state of \\NL/SPARQL benchmarks} for KGQA tasks. 
    \item We introduce \emph{Spider4SPARQL - the most complex to-date cross-domain benchmark dataset for KGQA} tasks, which contains 4,721 unique and novel SPARQL pairs and their corresponding 166 multi-domain knowledge graphs and ontologies. The new benchmark dataset and corresponding knowledge graphs and ontologies are made publicly available. In addition, we will submit Spider4SPARQL as a new benchmark dataset for KGQA tasks to the \emph{KGQA leaderboard}\cite{KGQA_leaderboard}\footnote{https://kgqa.github.io/leaderboard/}.
    \item Our experimental results with small, medium and large language models against Spider4SPARQL show that execution accuracies for translating natural language to SPARQL reach up to 45\%. The \emph{results demonstrate that the Spider4SPARQL benchmark is very challenging} and existing systems need substantial improvements to achieve execution accuracies of 90\% or above, as we see for less challenging benchmarks.
\end{itemize}

The rest of this paper is structured as follows. Section \ref{sec:related_work} describes related work and existing knowledge graphs as well as frequently used benchmark datasets for KBQA tasks. Section \ref{sec:Spider4SPARQL_Benchmark_details} describes the process of creating Spider4SPARQL, including the construction of the new knowledge graphs and the translation of the queries from SQL to SPARQL. Section \ref{sec:eval_quality} and  \ref{sec:characterstics_of_spider4sparql} analyze the complexity and the quality of Spider4SPARQL.  Section \ref{sec:experiments} shows a detailed  experimental evaluation of various state-of-the-art language models on the newly created benchmark. We conclude in Section \ref{sec:conclusion}. 

\section{Related Work and Analysis of Existing Benchmarks}
\label{sec:related_work}
In this section, we review the related work on the major knowledge graphs as well as the most commonly-used datasets for evaluating KGQA systems. 

\subsection{Knowledge Graphs}
The two largest and most widely used open source knowledge graphs for creating KGQA benchmarks are \emph{Wikidata\footnote{\url{https://www.wikidata.org/}}} and \emph{Dbpedia\footnote{\url{https://www.dbpedia.org/}}}. 

\emph{Wikidata} is the largest open source knowledge graph. ``Wikidata is a free, collaborative, multilingual, secondary database, collecting structured data to provide support for Wikipedia, Wikimedia Commons, the other wikis of the Wikimedia movement, and to anyone in the world.'' Similar to Wikipedia\footnote{\url{https://www.wikipedia.org/}}, Wikidata can be edited by anyone, i.e. the content in the knowledge graph can change from day to day and is also not guaranteed to be correct. Wikidata currently contains  102,834,755 data items and is available in 331 languages. 

\emph{DBpedia} is a curated knowledge graph with manually created specifications in the DBpedia Mappings Wiki\cite{dbpedia}. Each release of this ontology corresponds to a new release of the DBpedia dataset which contains instance data extracted from the different language versions of Wikipedia. DBpedia is maintained by the DBpedia community, which creates mappings from Wikipedia information representations to the DBpedia ontology. 

\subsection{Benchmark Datasets for Evaluating KGQA Systems}

In the previous paragraph, we introduced the key knowledge graphs used in KGQA benchmarks. We now discuss benchmark datasets that are based on these knowledge graphs. 

Note that since these knowledge graphs are constantly being updated, it frequently happens that a benchmark dataset that was developed at one point in time no longer corresponds to the most current version available. In other words, some of the originally developed NL/SPARQL pairs might no longer work due to updates in the underlying knowledge graphs.

\emph{The DBpedia Neural Question Answering (DBNQA) Dataset} is a large-scale dataset consisting of 894,499 NL/SPARQL pairs. The SPARQL queries in the DBQNA dataset were generated using 5,215 SPARQL query templates from two existing datasets LC-QuAD and QALD-7 train\footnote{\url{https://github.com/ag-sc/QALD/blob/master/7/data/qald-7-train-en-wikidata.json}}. The natural language questions of DBNQA were generated by replacing the existing entities in each NL question with new entities generated with the SPARQL templates. The LC-QuAD 1.0 dataset provides an intermediary natural language question where the SPARQL entities used in the corresponding query are tagged. The 215 natural language templates from QALD-7 were created by tagging the entities in the NL questions manually\footnote{https://github.com/AKSW/DBNQA}. Unlike the datasets described in the following paragraphs, the DBNQA dataset authors do not provide any explicit dataset partitioning, i.e. explicit train and test sets, which can lead to models memorizing template patterns\cite{dbqna_test_train_splits}.

\emph{QALD-9} is a multi-lingual dataset developed within the QALD challenge\footnote{https://qald.aksw.org/} which began in 2011 and continued over 8 years. It consists of 408 train questions developed in the challenges from previous years and 150 new test questions gathered from real-world system logs that are executable against the DBpedia knowledge graph. \cite{qald} The questions cover various complexities, including counts, superlatives, comparatives, and temporal aggregators. 

\emph{LC-QuAD 1.0} is a dataset derived from DBpedia that was automatically generated using 37 different SPARQL query templates and features 5,000 NL/SPARQL pairs (4K train, 1K test) \cite{lcquad1.0}. The NL questions in this benchmark were not manually generated but rather developed using an NL question template as an intermediary step. Finally, the questions were corrected with human intervention. 

This reveals one key issue with this dataset. Many of the queries contain the entire entity in the natural language question, meaning it is relatively easy for current neural networks to perform \emph{named entity recognition} (NER) with very high accuracy on the NL questions. 

Take the following example of an NL/SPARQL pair from the LC-QuAD 1.0 test set. 

\textbf{Example NL/SPARQL pair from LC-QuAD 1.0}: \emph{"Which \textbf{architect} of \textbf{Marine Corps Air Station Kaneohe Bay} was also \textbf{tenant} of \textbf{New Sanno hotel}?"}. 

The corresponding SPARQL query is as follows:

{\small
\begin{Verbatim}

SELECT DISTINCT ?uri WHERE { 
  dbr:Marine_Corps_Air_Station_Kaneohe_Bay
  dbp:architect ?uri. 
  dbr:New_Sanno_Hotel dbo:tenant ?uri
} 
\end{Verbatim}
}

The terms highlighted in the NL question are exactly the same as those in the underlined resource URIs in the SPARQL query. Performing NER on the NL question and matching the entities to the corresponding resources and properties to generate a SPARQL query is therefore relatively straightforward.

Questions written by human users tend to be more ambiguous and do not necessarily use the exact terms from the database schema or exact values from the database. Let us compare with the following example from the Spider4SPARQL dataset, which was manually generated. 

\textbf{Example NL/SPARQL pair from Spider4SPARQL}: \emph{"Give the \textbf{name} of the \textbf{nation} that uses the greatest amount of \textbf{languages}"}.

The corresponding SPARQL query is as follows:

{\small
\begin{Verbatim}
SELECT ?T1_name (count( *) as ?agg) 
WHERE {
  ?T1 a :country .
  ?T2 a :countrylanguage .
  ?T2 :countrylanguage#ref-countrycode ?T1 .  
  ?T1 :country#name ?T1_name . 
  ?T1 :country#code ?T1_code 
} 
GROUP BY ?T1_name  
ORDER BY DESC (?aggregation_all) 
LIMIT 1
\end{Verbatim}
}

Only one of the terms highlighted in the NL question exactly corresponds to a data property in the knowledge graph. Another weakness of the LC-QuAD 1.0 dataset is the template-generated SPARQL queries. The lack of diversity in the dataset has the potential to negatively impact neural network-based KGQA approaches.

\emph{Large-Scale Complex Question Answering Dataset 2.0} or LC-QuAD 2.0 is the second edition of the LC-QuAD benchmark \cite{lcquad2.0}. It comprises 30K individual question-answer pairs (24K train, 6K test), a 500\% increase compared to the first edition of the benchmark. Due to the increase in size and complexity compared to the first edition of the benchmark, LC-QuAD 2.0 is currently the benchmark of choice in the majority of recent KGQA systems papers. 

Similar to the first edition of the benchmark, the SPARQL queries in this dataset were also curated using 22 unique SPARQL query templates (15 fewer than the first edition). The natural language questions for each query were developed via the Amazon Mechanical Turk crowdsourcing platform.
Unlike LC-QuAD 1.0, the second edition of the dataset is executable against both major open source knowledge graphs, \emph{Wikidata} and \emph{DBpedia}. While systems achieved accuracies of 85\% \cite{2019-SubQG} as early as 2019 on its predecessor LC-QuAD 1.0, they did not achieve similar accuracies on LC-QuAD 2.0 until 2021 from systems leveraging transformer architectures \cite{modern_baselines}.

The creators of this dataset do not provide the dataset analysis on the difficulty of the dataset solely in terms of the difficulty of the SPARQL queries but also linguistic characteristics in the NL questions. 

\section{Spider4SPARQL - A New  Benchmark for Evaluating Text-to-SQL Systems}
\label{sec:Spider4SPARQL_Benchmark_details}
In this section, we describe how we built our novel benchmark dataset Spider4SPARQL. 

First, we describe how we \emph{automatically construct knowledge graphs} given a set of 166 relational databases (see Section \ref{sec:constructing_KGs}). Next, we explain how we \emph{automatically generate SPARQL queries} based on the constructed knowledge graphs (see Section \ref{sec:converting_sql_to_sparql}). 

\subsection{Constructing Knowledge Graphs based on Relational Databases}
\label{sec:constructing_KGs}

Our starting point is the Spider dataset, built for querying Text-to-SQL systems based on relational databases. The Spider dataset is a great starting point for developing a new KGQA benchmark because of its wide-ranging query types, manually generated natural language questions and rich and diverse data content, which covers 138 different domains from data about flights and airlines to orchestras and concerts. 
%\pcm{say in a few words why Spider is a great starting point}
The goal of this paper is to build a KGQA benchmark for querying knowledge graphs. Hence, we need to translate the relational databases and the SQL queries into knowledge graphs and SPARQL queries, respectively. The main challenge is \emph{how to perform this translation efficiently and effectively} given that the relational databases and the SQL queries of the Spider dataset adhere to a far less strict formalism than required for the methods normally used to convert relational databases to RDF-based knowledge graphs. 

In this section, we give a high-level overview of the methods used to construct knowledge graphs (KGs) from an existing relational database (DB).

\subsubsection{Applying Direct Mapping between DBs and KGs}

The W3C provides two (not fully compatible) standards for mapping relational sources to RDF graphs: \emph{Direct Mapping}~\cite{W3Crec-RDB-Direct-Mapping} and \emph{R2RML}~\cite{W3Crec-R2RML}. The first, Direct Mapping, is a W3C recommendation stipulating how to transform a DB into a KG while preserving the information and the vocabulary used in the DB. Starting from a database schema: each \textit{table} becomes a \textit{class} in the ontology, each \textit{attribute} becomes a \textit{(data) property} relating objects in the class to \emph{RDF literals} (e.g., strings, integers, etc.), and each \textit{foreign key relationship} is transformed into an (object) \textit{property} relating objects between the classes corresponding to the tables participating in the relationship. 

To build a KG based on an existing DB, we can apply the \textit{Direct Mapping} approach. The preserved vocabulary allows us to have a direct correspondence between the DB and the produced KG. This puts us in an advantageous setting where we can directly apply well-established techniques developed for SQL to the SPARQL case. 

A major difference between a DB and a KG created through Direct Mapping is the fact that the latter does not include a \emph{schema}. Hence, all the information present in the \textit{DB schema is essentially lost} during the transformation to the KG. This behavior of Direct Mapping is due to the recommendation being specifically designed to be agnostic of any ontology language that might be adopted for the constructed KG.

\begin{figure}[h]
  \includegraphics[trim=0cm 0cm 0cm 1cm, width=\linewidth]{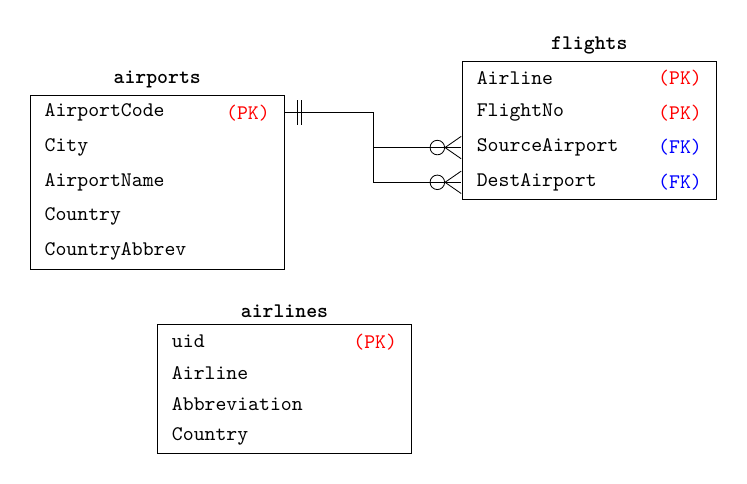}
  \caption{Schema of the \texttt{flight\_2} database from the Spider dataset. Note that there are no primary/foreign keys between the tables \texttt{flights} and \texttt{airlines}.} 

  \label{figure:running-ex-schema}
\end{figure}

Let us illustrate the concept of Direct Mapping with the relational database \texttt{flights\_2} from the Spider dataset shown in Figure \ref{figure:running-ex-schema}. Note that this database has a missing primary/foreign key between the tables \texttt{flights} and \texttt{airlines}.

\begin{figure}[H]
  \centering
  \includegraphics[width=\linewidth]{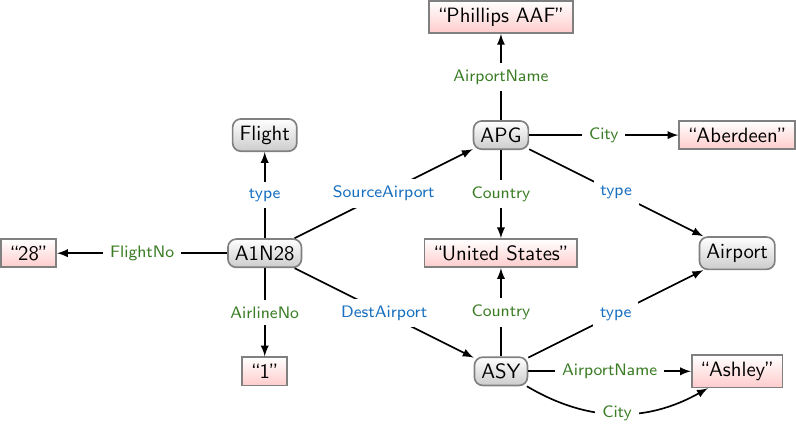}
  \caption{\label{figure:running-ex-kg} A portion of a knowledge graph showing the source and destination airports of flights. Classes are highlighted in grey, data property edges are labelled in green and object property edges in blue. Literals correspond to the base data of a relational database and are shown in red.}
\end{figure}

An example of Direct Mapping for a portion of the \texttt{flights\_2} database is displayed in Figure~\ref{figure:running-ex-kg}. We can see that the original tables "flights" and "airports" have been converted into the \emph{classes} "Flight" and "Airport", respectively. Also note that the attributes "FlightNo" and "Country" have been converted into \emph{data properties}. Finally, the primary/foreign key relationships between "airports" and "flights" have been converted to \emph{object properties}. 

One drawback of Direct Mapping is that it requires the source schema to be sufficiently structured (i.e., primary and foreign key constraints must be explicitly declared). However, as shown in Figure \ref{figure:running-ex-schema}, the schema is missing a primary/foreign key relationship between the tables "flights" and "airports".  Hence, Direct Mapping does not work in that scenario.

\emph{R2RML} is a W3C recommendation for mappings that overcomes this limitation. Through R2RML, users can specify mappings by hand. This allows for greater flexibility because the data source can be mapped to any ontology, regardless of the vocabulary and the structure of the data source. However, in the \emph{Virtual Knowledge Graph (VKG) } (described below), this comes at the price of sacrificing performance since the knowledge graph is not materialized (physically stored in a triple store) but derived through mappings during query processing. 

\begin{figure}[h!]
  \centering
  \includegraphics[width=.9\linewidth]{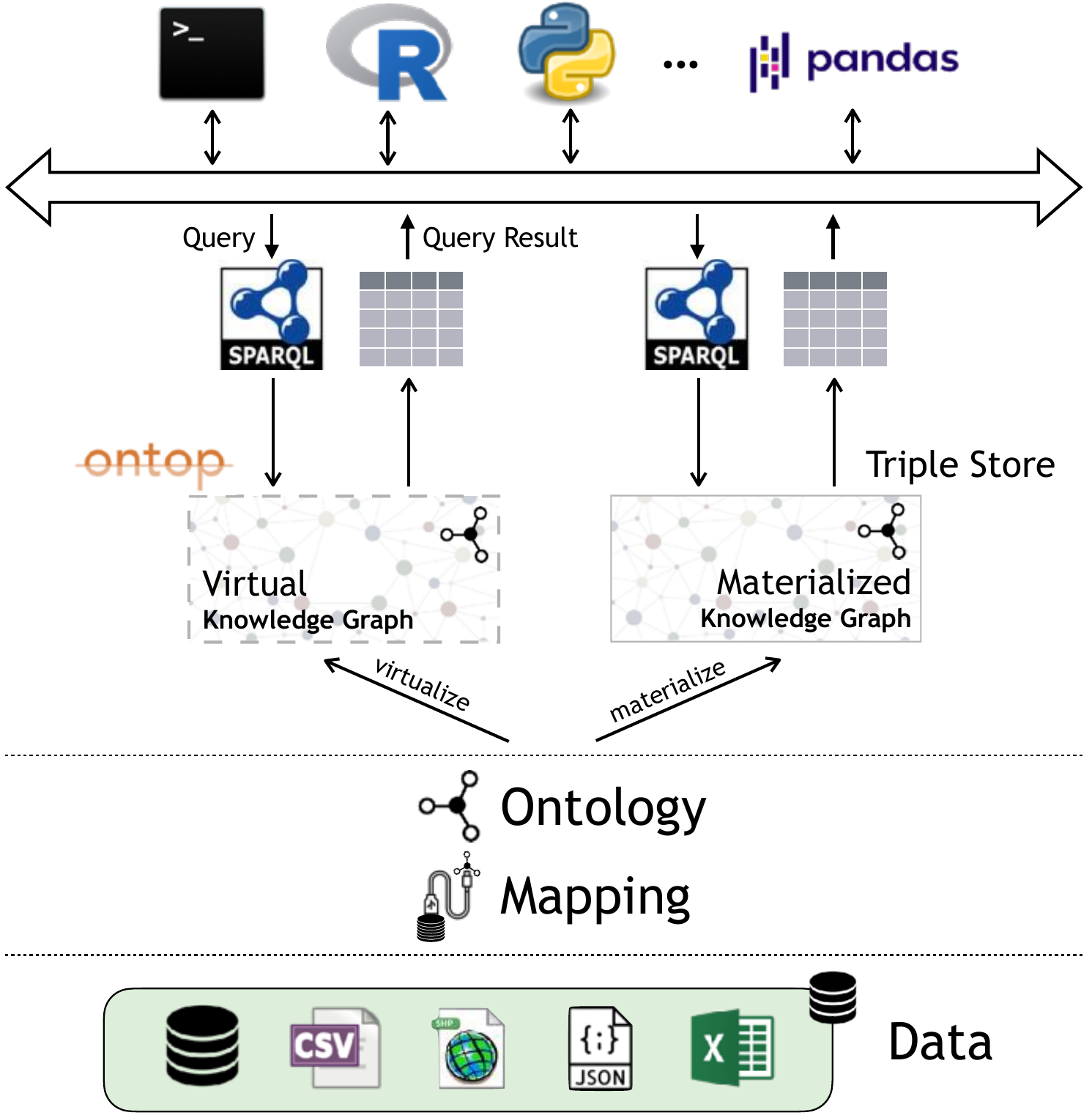}
  \caption{Conceptual framework of virtual knowledge graphs. Various data sources are exposed via ontology mappings either as a virtual or a materialized knowledge graph.}
  \label{fig:conceptual-framework}
\end{figure}

An example of a \emph{Virtual Knowledge Graph} is depicted in Figure~\ref{fig:conceptual-framework}. In this approach, a legacy (relational) data source is exposed as a \emph{virtual} RDF graph via, e.g. Direct Mapping. The KG is virtual in the sense that RDF triples are not stored anywhere, but rather SPARQL queries are translated on-the-fly into equivalent SQL queries that are executed against the data source directly. To convert the original relational databases of the Spider dataset into knowledge graphs, we use the VKG-system Ontop \cite{CCKK*17}.

\subsubsection{Converting SQLite Databases to PostgreSQL}
The original Spider dataset was provided as a series of SQLite databases. However, since Ontop does not support SQLite, the original Spider databases were converted from SQLite to PostgreSQL. 
 
When converting from SQLight to Postgres, we faced two main challenges. 

\begin{itemize}
\item \emph{Challenge 1 - Data Modeling Errors}: Many of the Spider databases are missing primary and foreign key relationships (as also shown in Figure \ref{figure:running-ex-schema}). In order to translate the SQLite databases to PostgreSQL, it was necessary to add these keys, which resulted in an overall cleaner and properly modelled dataset.
\item \emph{Challenge 2 - Data Type Errors}: In addition to data modeling errors, there were many data type errors in the SQLite version of the Spider databases as well. For example, many primary/foreign key relationships had different data types, i.e. an \emph{int} PK and \emph{text} FK. There were also many columns with numerical data that were created with a text datatype. We addressed these issues while translating from SQLite to PostgreSQL.
\end{itemize}

\subsection{Converting SQL Queries to SPARQL}
\label{sec:converting_sql_to_sparql}

In this section, we describe how we convert SQL queries to SPARQL. The major challenge in this context is \emph{how to perform this approach automatically} given our previously constructed KGs.

\subsubsection{Converting NL to SemQL}
The first step in converting SQL to SPARQL is translating the SQL queries to an intermediary language called SemQL. SemQL was first introduced as a context-free grammar to represent SQL queries in IRNet, one of the first systems submissions to the Spider Leaderboard \cite{IRNet}. The advantage of using SemQL as an intermediate language is that it enables translating an NL to any query language, such as SQL or SPARQL.

Table \ref{table:SemQL_2_0} depicts all of the SQL operations that are included in the SemQL grammar, such as select, project, and join. The sections highlighted in green required extensions for translating SemQL to SPARQL - as discussed in the next section.

Group By and Having clauses are included in the deterministic translation of SemQL to SPARQL. 

\begin{table}[!ht]
\setlength{\tabcolsep}{2pt}
  \caption{SemQL Grammar: An intermediate language for translation between a natural language question and a query language.}
  {\tt
    \begin{tabular}{lcl}
      Z & ::= & \colorbox{green!20}{intersect R R} $\mid$ union R R $\mid$\\
        && \colorbox{green!20}{except R R $\mid$ R}\\
      R & ::= & Select $\mid$ Select Filter $\mid$ \\
        &&Select Order $\mid$ Select Superlative $\mid$\\ 
        &&Select Order Filter $\mid$\\ 
        &&Select Superlative Filter \\
      Select & ::= & distinct N $\mid$ N \\
      N & ::= & A $\mid$ A A $\mid$ A A A $\mid$ A A A A  $\mid$ A A A A A \\
      Order & ::= & asc A $\mid$ desc A \\
      Superlative & ::= & most V A $\mid$ least V A \\
      Filter & ::= & and Filter Filter$\mid$ or Filter Filter $\mid$ \\
        &&= A V $\mid$ = A R $\mid$ $\neq$ A V $\mid$ $\neq$ A R $\mid$ \\
        &&$<$ A V $\mid$ $<$ A R $\mid$ $>$ A V $\mid$ $>$ A R $\mid$ \\
        &&$\leq$ A V $\mid$ $\leq$ A R $\mid$ $\geq$ A V $\mid$ $\geq$ A R $\mid$\\
        && \colorbox{green!20}{between A V V $\mid$ between A R}\\
        && A R $\mid$ not in A R \\
      A & ::= & max Op $\mid$ min Op $\mid$ count Op $\mid$\\
        && sum Op | avg Op $\mid$ Op \\
      Op & ::= & - C T C T  $\mid$ + C T C T  $\mid$ * C T C T  $\mid$ \\ &&  $\div$ C T C T  $\mid$  C T C T \\
      C & ::= & column\\
      T & ::= & table\\
      V & ::= & value\\
    \end{tabular}
  }
  \label{table:SemQL_2_0}
\end{table}

\subsubsection{Converting SemQL to SPARQL}

The second step of our conversion process is translating SemQL into SPARQL. Using a deterministic translation from SemQL to SPARQL we were able to successfully translate the majority, i.e. 95\% of the queries automatically. About 5\% of the queries needed to be corrected manually. This is due to some SQL operations present in the original Spider dataset and which are not covered by SemQL, such as \textit{LIMIT} clauses with values greater than one, e.g. LIMIT 3. 

\subsubsection{Challenges of SQL to SPARQL Conversion}
We now discuss the challenges we faced when converting \emph{aggregation queries} and \emph{set operation queries} from SQL to SPARQL via SemQL.

\paragraph{Aggregation Queries} \label{aggregation_queries}

The following example shows that SQLite (the DB engine used in the original Spider dataset) allows queries with aggregated and non-aggregated variables in the query projection. For instance, the attribute \texttt{t1.City}  is part of the projection but not in the \texttt{GROUP BY} clause:

{\small
\begin{verbatim}
SELECT t1.City, count(*)
FROM airports AS t1
JOIN flights as t2
ON t1.AirportCode = t2.SourceAirport
GROUP BY t2.SourceAirport
\end{verbatim}
}

Note that the query above is not compliant with the SQL standard. Likewise, SPARQL cannot execute this type of query as is and returns a {\tt MALFORMED QUERY} error. According to the SPARQL 1.1 Recommendation, queries that have aggregates in their projection may only include non-aggregated variables when these are included in a GROUP BY statement\footnote{https://www.w3.org/TR/sparql11-query/\#groupby}. 

Therefore, in order to execute this type of query in SPARQL,
the query must be translated as follows: 
\begin{small}
\begin{verbatim}
SELECT ?city (count(*) as ?agg)
WHERE { 
  ?t1 a :airports . 
  ?t2 a :flights . 
  ?t2 :flights#ref-sourceairport ?t1 .
  ?t1 :airports#city ?city.
  ?t2 :flights#sourceairport ?sourceairport 
}   
GROUP BY ?city ?sourceairport
\end{verbatim}
\end{small}
The query above is now executable because the variables in the projection are now included in the GROUP BY statement.

\paragraph{Set Operation Queries}\label{set-op-queries}

Certain SQL keywords do not exist in SPARQL, such as \texttt{INTERSECT} and \texttt{EXCEPT}. In order to achieve these kinds of set operations in SPARQL, alternative formulations are required. Consider the following SQL example taken from the Spider dataset:

{\small
\begin{verbatim}
SELECT City
FROM airports
WHERE AirportCode = 'MMI'
INTERSECT
SELECT City
FROM airports
WHERE AirportCode = 'AHN'
\end{verbatim}
}

The corresponding SPARQL query is as follows:

\begin{small}
\begin{verbatim}
SELECT ?t1_city WHERE { 
  ?t1 a :airports . 
  ?t1 :airports#city ?t1_city . 
  ?t1 :airports#airportcode ?airportcode . 
  FILTER(?t1_airportcode = 'MMI') . 
  FILTER(?t1_city IN (?t2_city)) .
  ?t2 a :airports . 
  ?t2 :airports#city ?t2_city . 
  ?t2 :airports#airportcode ?airportcode . 
  FILTER(?airportcode = 'AHN') . 
}
\end{verbatim}
\end{small}
The SQL-\texttt{INTERSECT} construct can be replicated in SPARQL by adding a \texttt{FILTER-IN}-clause for the variables that mirror the projected columns in the \texttt{SELECT} statements in the SQL query. This clause checks if the results from the variable \texttt{?t1\_city} are also in the results from the variable \texttt{?t2\_city}.

\subsection{SPARQL Query Translation Quality Control}
\label{sec:eval_quality}

To ensure the quality of the translations from SQL to SPARQL in the Spider4SPARQL benchmark, we use \emph{execution accuracy}. We compare the results sets of the newly created SPARQL queries (executed over our knowledge graphs) against the result sets of the SQL queries from the Spider dataset. %We use  to determine if the SPARQL queries have been correctly translated. In other words, we compare the result sets of the SQL queries with those of the SPARQL queries.

We executed the 4,721 Spider SQL queries of the train set and the 540 queries of the test set. We also executed the corresponding SPARQL queries against the newly created knowledge graphs. Table \ref{tab:dev_train_query_count} shows the number of \emph{unique} queries in the dataset along with the execution accuracy between the results of the SQL and the respective SPARQL queries. The results show that just 6\% of the test set and 6\% of the train set could not be translated accurately. The errors in the translation were due to incorrect SemQL translations, i.e. the SemQL representation did not select the correct properties for connecting classes. The 6\% of queries from both the test and train set that could not be translated correctly with SemQL were manually translated. 

\begin{table}[h!]
    \centering
    \caption{The table shows the number of unique SQL and SPARQL queries of the train and test set, respectively. The table also shows the execution accuracy of the SQL and corresponding SPARQL queries.}
    \begin{tabular}{l|r|r|r}
        \hline
        Dataset Split & SQL & SPARQL & Execution Accuracy of \\ & & & Auto. Translated Queries \\
        \hline
        \hline
        Train set & 4,721 & 4,721 & 94\% \\
        Test set & 540 & 540 & 94\% \\
        \hline
    Total & 5,235 & 5,235
    \end{tabular}

    \label{tab:dev_train_query_count}
\end{table}

\section{Dataset Analysis}
\label{sec:characterstics_of_spider4sparql}

We now analyze the complexity of our new benchmark dataset Spider4SPARQL. 

\paragraph{Natural Language Question Generation} 

LC-QuAD 1.0 and DBNQA use a template-based automated method for generating natural language questions and applying paraphrasing. LC-QuAD 2.0 relies on crowdsourcing to correct paraphrased questions. The questions in the QALD-9 dataset are manually generated.

Spider4SPARQL provides a different approach to the automatic and semi-automatic (LC-QuAD 2.0) approaches, in that it contains 9,693 natural language questions that were manually generated for the Spider dataset. Manually generated questions are generally not susceptible to the pitfalls that often befall automatically and semi-automatically generated questions: over-fitting and only covering entities generated in the template-based approaches.

Table \ref{tab:query_complexity} shows the complexity of the queries in our new benchmark compared to existing benchmarks. In order to perform a systematic comparison, we group the queries into \emph{three different categories}, namely single-hop queries, multi-hop queries and aggregation queries\footnote{In SQL-terminology, the terms single-hop and multi-hop queries would be single-table and multi-table queries, respectively.}. The categorization is based on the SQL concept of SPJA-queries, i.e. select-project-join-aggregate.

\begin{table*}
\centering
\caption{Complexity of the NL questions and SPARQL queries of various NL/SPARQL benchmark datasets compared with our new benchmark dataset Spider4SPARQL.}
\label{tab:query_complexity}
\begin{tabular}[h!]{l|r|r|r|r|r}
\hline
\hline
Dataset Features & LC-QuAD 1.0 & LC-QuAD 2.0 & DBNQA & QALD-9& \textbf{Spider4SPARQL}\\ 
\hline
\hline
\#NL questions & 5,000 & 30,226 & 894,499 & 558 & 9,693\tablefootnote{There are 
8,659 NL questions in the train set and 1034 NL questions in the test set.} \\ 
Manually generated NL questions  & - & (+) & - & + & + \\
\#SPARQL queries train & 4000 & 24,180&894,499\tablefootnote{No explicit partitionings were provided by the dataset authors}& 408 & 4,721\tablefootnote{In the original Spider dataset there can be multiple unique NL questions per SQL query .}\\
\#SPARQL queries test & 1,000 & 6,046 &  & 150 & 540 \\ 

\hline

\textbf{Single-hop queries} & & & & & \\
\hline
Max. \#selections (filters) & - & 1 & 1 & 2 & 2\\
Max. \#projections & 1 & up to 2 & 1 & 1 & up to 6 \\ 
\# comparison operators & 1 & 1 & 2 & 7 & 7\\
Order by (ASC/DESC) & - & + & + & + & + \\
Mathematical operations & - & - & - & + & + \\
\hline

\textbf{Multi-hop queries} & & & & & \\
\hline
Set operations & - & - & (+) & + & + \\
Max. \#triple patterns & 2 & 3 & 3 & 5 & 10\\
Subqueries & - & - & - & - & 5 \\
\hline

\textbf{Aggregation queries} & & & & & \\
\hline
Types of aggregations & count & count & count & count & count, min, max, avg, sum \\
Group by/having clauses & - & - & - & + & + \\
Max. \#aggregations & 1 & 1 & 2 & 1 & 3\\
\hline
\end{tabular}
\end{table*}

\paragraph{Single-hop queries}
Spider4SPARQL contains the highest complexity in terms of single-hop queries, especially in the number of projected variables in the select statement. 

Although QALD-9 contains queries with a variety of comparison operators, these queries comprise just 5\% (31 queries) of the total number of queries in both its train and test sets.  

QALD-9 is also the only other benchmark that contains mathematical operations, however, there are no queries with mathematical operations in the train set, and just 2 queries with mathematical operations in the test set. The Spider4SPARQL benchmark contains multiple queries with mathematical operations in both the train and test sets.  

\paragraph{Multi-hop queries}

As discussed in Section \ref{sec:converting_sql_to_sparql}, Spider4SPARQL contains queries with various set operations and as many as 6 hops in a query. Spider4SPARQL also contains many queries with subqueries in the filter. 
The DBNQA and QALD-9 datasets only support union queries and have a maximum of 3 hops per query.

The following is an example of a more complex query from Spider4SPARQL which contains multiple hops, aggregations, a filter, a subquery and a group by clause. 

\textbf{Example NL/SPARQL pair from Spider4SPARQL}: \emph{”Find the number of cities in each district whose population is greater than the average population of cities?”}
 
\begin{table}
\centering
\caption{The table shows the execution accuracy of the SPARQL queries produced by 3 trained NL-to-SPARQL models executed against the 1,034 ground truth samples from the Spider4SPARQL test set.}
\label{tab:trained_model_results}
\begin{tabular}[h!]{l | l l l }
\hline
\toprule
Model & T5-Small& ValueNet4SPARQL&  T5-Base \\
\#Parameters    & 60M & 148M  & 220M \\
\hline
Execution Accuracy & 27\% &41\%  &42\%\\
\hline
\end{tabular}
\end{table}
\begin{small}
\begin{verbatim}
SELECT (count(*) as ?agg) ?t1_district 
WHERE { 
  ?t1 a :city . 
  ?t1 :city#district ?t1_district . 
  ?t1 :city#population ?t1_pop . 
  FILTER(?t1_pop > ?agg_t22) .       
  { SELECT (avg(?t22_pop) as ?agg_t22) 
    WHERE {
      ?t22 a :city . 
      ?t22 :city#population ?t22_pop .       
    }
  }
} GROUP BY ?t1_district
\end{verbatim}
\end{small}

\paragraph{Aggregation queries} Spider4SPARQL supports the largest number of aggregations out of all of the datasets. The other benchmark datasets only support the count aggregation.  It also contains queries with up to 3 aggregations in a single query.

The following example shows a query containing the maximum number of aggregations per query in the dataset:
\begin{small}
    \begin{verbatim}
SELECT (avg(?age) as ?avg ) 
  (min(?age) as ?min ) 
  (max(?age) as ?max ) 
WHERE { 
  ?t1 a :singer . 
  ?t1 :singer#age ?age . 
  ?t1 :singer#country ?country . 
  FILTER(?country = 'France') 
}
\end{verbatim}

\end{small}

\section{Experimental Evaluation}
\label{sec:experiments}

\begin{table}
\centering
\caption{The table shows the execution accuracy of the SPARQL queries produced by GPT-3.5 using zero and few-shot learning. The few-shot accuracy represents the results of the 10-fold few-shot experiment.}
\label{tab:gpt_results}
\begin{tabular}[h!]{ l | l}
\hline
\toprule
Model & GPT-3.5 \\ 
\#Parameters & 175B \\

\hline

Execution Accuracy (zero-shot) & 8\% \\ 
Execution Accuracy (10 shots) &  45\% ($\pm$4.41\%.) \\
\hline

\end{tabular}
\end{table}

To demonstrate the complexity of the Spider4SPARQL benchmark, we evaluate 4 state-of-the-art language models on the newly created benchmark. In particular, we evaluate 3 \emph{small and medium-sized language models} that are fine-tuned using our train set. We also evaluate 1 \emph{large language model} in a zero- and few-shot learning setting using samples from our train set.
\subsection{Experimental Setup}
\label{sec:experimental_setup}
\emph{Hardware}: All fine tuning experiments were executed on a single Tesla V100
GPU (32GB memory) with an Intel(R) Xeon(R) CPU E5-2650
v4 (4 cores) and 16GB memory.
\emph{Implementation}:
The model checkpoints have been chosen based on the top execution accuracy on the test set.

ValueNet4SPARQL was trained for 100 epochs with a batch size of 8. It reached top execution accuracy around epoch 40. 

T5-Small and T5-Base were trained for 28,000 steps, with a batch size of 5. T5-Small reached top execution accuracy around step 17,000 and T5-Base around step 14,000. 

For the zero- and few-shot experiments with GPT-3.5, to encourage deterministic and highly focused output from the model, we used a temperature setting of 0 and set the nucleus sampling to 1.
To reproduce our experiments we release all code (including hyperparameters for each model), prompts, and the benchmark dataset at the aforementioned Github repository in Section \ref{sec:intro}.

\subsection{Baseline Models}
\label{sec:baseline_experiments}

\paragraph{ValueNet4SPARQL}
ValueNet4SPARQL is an adapted version of ValueNet\cite{brunner2021valuenet} that enables translating natural language questions to SPARQL queries. ValueNet is based on a small language model which has 148 million parameters. It reuses some of the components of an earlier Text-to-SQL system IRnet\cite{IRNet}. Its novelty lies in that it incorporates a method to extract values from natural language questions and infer possible value candidates from the base data of a given database, even if the values are not explicitly stated in the natural language question. The system consists of a BART encoder and a grammar-based decoder to translate natural language questions into the context-free grammar SemQL, which is then deterministically translated into SQL or SPARQL.  

\paragraph{T5-Small \& T5-Base}
T5-Small and T5-Base are two language models from the T5 family of Text-to-Text language models \cite{Text-to-text-transformer}. These can be considered small-sized language models with 60 million and 220 million parameters, respectively. We have chosen these T5 models for our experiments since they are currently the highest performing models \cite{modern_baselines} on several KGQA benchmark datasets, especially the Large-scale Complex Question Answering datasets\footnote{https://github.com/KGQA/leaderboard/blob/gh-pages/dbpedia/lcquad.md}.

\paragraph{GPT-3.5}
We use GPT-3.5 turbo\footnote{Per the OpenAI website the performance of gpt-3.5-turbo is on par with Instruct Davinci and more affordable at \$0.002 / 1K tokens.} from OpenAI, as a few shot KGQA to showcase the complexity and novelty of our Spider4SPARQL benchmark in a zero- and 10-shot setting. 

\subsection{Analysis of Fine-Tuned Models}

ValueNet4SPARQL, T5-Small and T5-Base were trained using the full train and test set of the Spider4SPARQL benchmark. The inputs to each model include the natural language question, the SPARQL query, and its corresponding ontology. The execution accuracy of each model is summarized in Table \ref{tab:trained_model_results}. 

We can observe that T5-Base, which has 220M parameters, outperforms T5-Small, which has 60M parameters, by 15\%. This suggests that the model with a higher number of parameters does indeed have an advantage over a model with a lower number of parameters when applied to our benchmark dataset. These conclusions seem to be obvious. However, experiments executed in previous work\cite{modern_baselines} on the LC-QuAD datasets showed that T5-Small and T5-Base perform similarly. This suggests that these previous datasets might not be complex enough for more powerful language models to show its full strength.    

\begin{figure*}[h!t]
  \centering
  \includegraphics[width=\textwidth, keepaspectratio]{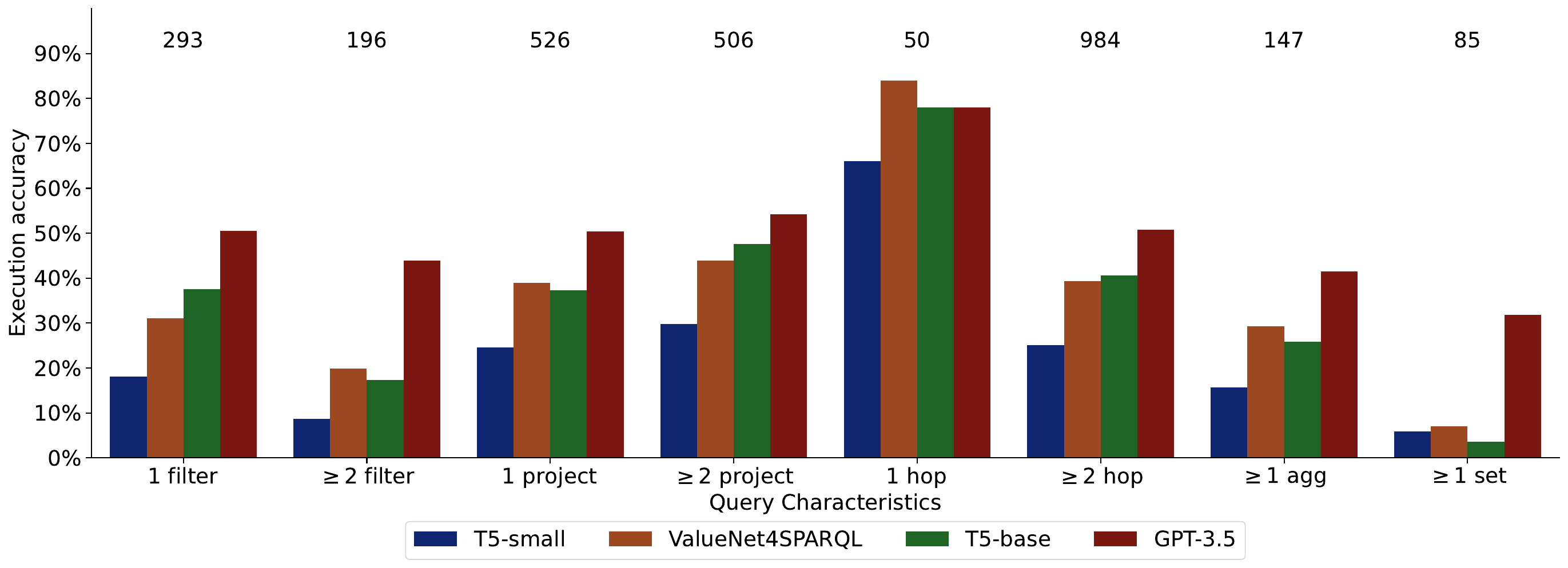} 
  \caption{The figure shows the execution accuracy for different query categories. The numbers on top of the bars show the number of sample queries per category.}
  \label{fig:query_characteristics}
\end{figure*}

ValueNet4SPARQL outperforms T5-Small by 14\% in execution accuracy and across the board in terms of query hardness in  Figure \ref{fig:hardness} and query characteristics in Figure \ref{fig:query_characteristics}. ValueNet4SPARQL has pre- and post- processing steps in its Text-to-SPARQL pipeline which help boost the systems's accuracy. These pre- and post-processing steps of the system show their impact in that ValueNet4SPARQL outperforms T5-Base, which has 49\% more parameters than ValueNet4SPARQL, in the hard and extra hard queries. It specifically performs better in queries that have more than one aggregation and more than one set operation. This is due to the post-processing step which guarantees compliance with the intricate and complex SPARQL syntax.

\subsection{Analysis of Zero- and Few-Shot Learning}

For the few-shot experiments, we selected a random sample of 100 unique natural language/SPARQL pairs from the train set. These 100 queries were split into 10 sets for a total of 10 examples per set. We evaluated GPT-3.5 10 times over the test set of 1,034 natural language questions using each aforementioned 10 query set as few-shot training data. 

The prompts for GPT3.5 include the \emph{natural language question}, the \emph{prefix of the knowledge graph} and the \emph{ontology of the knowledge graph}. We include the ontology in the prompt to inform GPT-3.5 about the structure and content of the knowledge graph that the query will be executed against. 

Assume that we want to translate the following simple natural language question \emph{"How many airports do we have?"} into SPARQL and execute it against the knowledge graph we have previously shown in Figure \ref{figure:running-ex-kg}.  Ideally, GPT-3.5 would be able to match the entities from the NL question, i.e. \emph{How many}, which indicates that the resulting query should include a \emph{COUNT} aggregation, and \emph{Airport} which corresponds to a class in the knowledge graph. 
A reduced prompt is shown in the example below\footnote{The full prompts and results are available in our GitHub repository referenced in Section\ref{sec:experimental_setup} }:

{\small 
\begin{verbatim}
Translate the following question into SPARQL: 
How many airports do we have? 

The prefix for the queries is: 
PREFIX : <http://valuenet/ontop/>

Use the following ontology:
'classes': ['airlines', airports', 'flights']

'object_properties': [flights#ref-DestAirport', 
'flights#ref-SourceAirport', 
'flights#ref-Airline']

'data_properties': ['airlines#Abbreviation', 
'airlines#Airline', 'airports#AirportCode', 
'airports#AirportName', 'flights#Airline', 
'flights#DestAirport']
\end{verbatim}
}

To evaluate the accuracy of the GPT-3.5 generated SPARQL queries, we again use the \emph{execution accuracy}, i.e. we compare the output of the generated SPARQL query to the output of the ground truth SPARQL query.

\paragraph{Zero-shot Results}
Table \ref{tab:gpt_results} shows that GPT-3.5 answered 8\% of the NL questions correctly. The majority of these queries contained a \emph{COUNT} aggregation and questions where the named entities matched the names of the required classes or properties.

GPT-3.5 could not produce runnable SPARQL queries for 25\% of the NL questions in the zero-shot scenario due to syntax errors. The majority of errors were due to properties being predicted as classes as well as missing relationships between classes, which demonstrates a misunderstanding of the structure of the ontology, even though it was provided in the prompt. Other errors include incorrectly predicted aggregations, filters on incorrect columns, missing having clauses, regex hallucinations and incorrect set operations.

\paragraph{Few-Shot Results}

GPT-3.5 reached an execution accuracy of 45\% (see the bottom line of Table \ref{tab:gpt_results}) in the few-shot scenario. These results demonstrate that giving the model just 10 different examples in the few-shot scenario leads to a significant performance increase of 37\% ($\pm$4.41\%) over the zero-shot case.

In the few-shot scenario, GPT-3.5 was unable to generate runnable SPARQL queries for 7\% of the queries on average due to syntax errors similar to those described in the zero-shot scenario. 

It is possible that the results from the few-shot experiments could be improved with advanced prompt engineering or manually selected examples, however, this is out of the scope of this benchmark paper and is left as a research challenge for further work.

\subsection{Analysis of Model Performance Across Query Complexity and Characteristics }

\begin{figure}
  \centering  \includegraphics[width=0.9\linewidth]{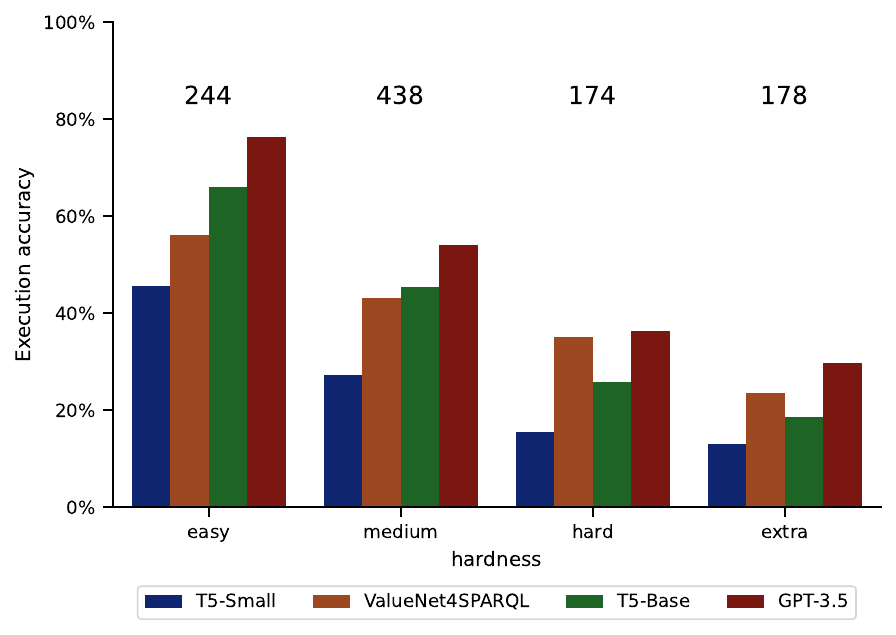} % Replace with your filename
  \caption{The figure shows the execution accuracy for easy, medium, hard and extra hard queries, across the baseline systems. The numbers on top of the bars indicate how many queries are in each category.}
  \label{fig:hardness}
\end{figure}

Let us now analyze the model performance across different types of queries. Figure \ref{fig:hardness} shows the performance of each model based on a query difficulty metric designed by the creators of the original Spider dataset\footnote{https://github.com/taoyds/spider/blob/master/evaluation\_examples/ README.md}. As shown in Figure \ref{fig:hardness}, the models performed best on "easy" type queries and worst on "extra" hard type queries. 

Finally, let us analyze how the models perform on the query categories that we previously defined in Table \ref{tab:query_complexity}. As we can see in Figure \ref{fig:query_characteristics}, all of the models perform well on simple queries with a single hop. We also observe that all models struggle with queries that have more than 1 set operator.

These results suggest that the models need further improvement to handle complex queries and they are far from being ready for real-life scenarios.

\section{Conclusion}
\label{sec:conclusion}
Existing state-of-the-art KGQA systems perform at 91\% for LC-QuAD 1.0 and 92\% for LC-QuAD 2.0\footnote{https://kgqa.github.io/leaderboard/}. This might give the impression that translating natural language to SPARQL is basically a solved problem. However, these datasets do not fully exploit the richness of the SPARQL query language.  

Against this background, we introduce \emph{Spider4SPARQL}  - the most complex, cross-domain dataset for evaluating KGQA systems to date. Our experiments with state-of-the-art language models of various sizes show that the dataset poses a significant challenge for the systems which only reach up to 45\% execution accuracy\footnote{The F1 score is the same as the execution accuracy because the precision and recall are equal. See \cite{nadig-etal-2020-database} for further evaluation metric explanations.} - which is far below the 92\% on existing benchmarks of lower complexity. We want to emphasize that the main contribution of the paper is not to evaluate existing LLMs like GPT-3.5 but to show that our novel benchmark dataset Spider4SPARQL is indeed very challenging. We believe that this benchmark can trigger new research efforts to evaluate and improve new KGQA systems.

In the future, we would like to expand the dataset to include more SPARQL-specific query forms like \emph{ASK}, \emph{DESCRIBE} and graph pattern filters like \emph{OPTIONAL} as well as making the benchmark multilingual. 

\section*{Acknowledgment}
This project has received funding from the European Union’s Horizon 2020 research and innovation program under grant agreement No 863410. 
We also thank Nico Kappeler and Davide Lanti for valuable discussions.

\bibliographystyle{IEEEtran}
\bibliography{main}

\begin{thebibliography}{10}
\providecommand{\url}[1]{#1}
\csname url@samestyle\endcsname
\providecommand{\newblock}{\relax}
\providecommand{\bibinfo}[2]{#2}
\providecommand{\BIBentrySTDinterwordspacing}{\spaceskip=0pt\relax}
\providecommand{\BIBentryALTinterwordstretchfactor}{4}
\providecommand{\BIBentryALTinterwordspacing}{\spaceskip=\fontdimen2\font plus
\BIBentryALTinterwordstretchfactor\fontdimen3\font minus \fontdimen4\font\relax}
\providecommand{\BIBforeignlanguage}[2]{{%
\expandafter\ifx\csname l@#1\endcsname\relax
\typeout{** WARNING: IEEEtran.bst: No hyphenation pattern has been}%
\typeout{** loaded for the language `#1'. Using the pattern for}%
\typeout{** the default language instead.}%
\else
\language=\csname l@#1\endcsname
\fi
#2}}
\providecommand{\BIBdecl}{\relax}
\BIBdecl

\bibitem{copestake_jones_1990}
A.~Copestake and K.~S. Jones, ``Natural language interfaces to databases,'' \emph{The Knowledge Engineering Review}, vol.~5, no.~4, p. 225–249, 1990.

\bibitem{EarlyWork2}
\BIBentryALTinterwordspacing
I.~Androutsopoulos, G.~D. Ritchie, and P.~Thanisch, ``Natural language interfaces to databases - an introduction,'' \emph{CoRR}, vol. cmp-lg/9503016, 1995. [Online]. Available: \url{http://arxiv.org/abs/cmp-lg/9503016}
\BIBentrySTDinterwordspacing

\bibitem{EarlyWork3}
\BIBentryALTinterwordspacing
A.-M. Popescu, A.~Armanasu, O.~Etzioni, D.~Ko, and A.~Yates, ``Modern natural language interfaces to databases: Composing statistical parsing with semantic tractability,'' in \emph{Proceedings of the 20th International Conference on Computational Linguistics}, ser. COLING ’04.\hskip 1em plus 0.5em minus 0.4em\relax USA: Association for Computational Linguistics, 2004, p. 141–es. [Online]. Available: \url{https://doi.org/10.3115/1220355.1220376}
\BIBentrySTDinterwordspacing

\bibitem{affolter2019comparative}
K.~Affolter, K.~Stockinger, and A.~Bernstein, ``A comparative survey of recent natural language interfaces for databases,'' \emph{The VLDB Journal}, vol.~28, no.~5, pp. 793--819, 2019.

\bibitem{sima2021bio}
A.~C. Sima, T.~M. de~Farias, M.~Anisimova, C.~Dessimoz, M.~Robinson-Rechavi, E.~Zbinden, and K.~Stockinger, ``Bio-soda: Enabling natural language question answering over knowledge graphs without training data,'' \emph{33rd International Conference on Scientific and Statistical Database Management}, 2021.

\bibitem{brunner2021valuenet}
U.~Brunner and K.~Stockinger, ``Valuenet: A natural language-to-sql system that learns from database information,'' \emph{International Conference on Data Engineering (ICDE)}, 2021.

\bibitem{katsogiannis2021deep}
G.~Katsogiannis-Meimarakis and G.~Koutrika, ``A deep dive into deep learning approaches for text-to-sql systems,'' in \emph{Proceedings of the 2021 International Conference on Management of Data}, 2021, pp. 2846--2851.

\bibitem{WikiSQL}
\BIBentryALTinterwordspacing
V.~Zhong, C.~Xiong, and R.~Socher, ``Seq2sql: Generating structured queries from natural language using reinforcement learning,'' \emph{CoRR}, vol. abs/1709.00103, 2017. [Online]. Available: \url{http://arxiv.org/abs/1709.00103}
\BIBentrySTDinterwordspacing

\bibitem{lcquad1.0}
P.~Trivedi, G.~Maheshwari, M.~Dubey, and J.~Lehmann, ``Lc-quad: A corpus for complex question answering over knowledge graphs,'' in \emph{International Semantic Web Conference}.\hskip 1em plus 0.5em minus 0.4em\relax Springer, 2017, pp. 210--218.

\bibitem{lcquad2.0}
M.~Dubey, D.~Banerjee, A.~Abdelkawi, and J.~Lehmann, ``Lc-quad 2.0: A large dataset for complex question answering over wikidata and dbpedia,'' in \emph{Proceedings of the 18th International Semantic Web Conference (ISWC)}.\hskip 1em plus 0.5em minus 0.4em\relax Springer, 2019.

\bibitem{DBLP:journals/corr/abs-1906-09302}
\BIBentryALTinterwordspacing
X.~Yin, D.~Gromann, and S.~Rudolph, ``Neural machine translating from natural language to {SPARQL},'' \emph{CoRR}, vol. abs/1906.09302, 2019. [Online]. Available: \url{http://arxiv.org/abs/1906.09302}
\BIBentrySTDinterwordspacing

\bibitem{Spider}
\BIBentryALTinterwordspacing
T.~Yu, R.~Zhang, K.~Yang, M.~Yasunaga, D.~Wang, Z.~Li, J.~Ma, I.~Li, Q.~Yao, S.~Roman, Z.~Zhang, and D.~R. Radev, ``Spider: {A} large-scale human-labeled dataset for complex and cross-domain semantic parsing and text-to-sql task,'' \emph{CoRR}, vol. abs/1809.08887, 2018. [Online]. Available: \url{http://arxiv.org/abs/1809.08887}
\BIBentrySTDinterwordspacing

\bibitem{PLCD*08}
A.~Poggi, D.~Lembo, D.~Calvanese, G.~De~Giacomo, M.~Lenzerini, and R.~Rosati, ``Linking data to ontologies,'' vol.~10, pp. 133--173, 2008.

\bibitem{KGQA_leaderboard}
\BIBentryALTinterwordspacing
A.~Perevalov, X.~Yan, L.~Kovriguina, L.~Jiang, A.~Both, and R.~Usbeck, ``Knowledge graph question answering leaderboard: A community resource to prevent a replication crisis,'' in \emph{Proceedings of the Thirteenth Language Resources and Evaluation Conference}.\hskip 1em plus 0.5em minus 0.4em\relax Marseille, France: European Language Resources Association, Jun. 2022, pp. 2998--3007. [Online]. Available: \url{https://aclanthology.org/2022.lrec-1.321}
\BIBentrySTDinterwordspacing

\bibitem{dbpedia}
J.~Lehmann, R.~Isele, M.~Jakob, A.~Jentzsch, D.~Kontokostas, P.~Mendes, S.~Hellmann, M.~Morsey, P.~Van~Kleef, S.~Auer, and C.~Bizer, ``Dbpedia - a large-scale, multilingual knowledge base extracted from wikipedia,'' \emph{Semantic Web Journal}, vol.~6, 01 2014.

\bibitem{dbqna_test_train_splits}
\BIBentryALTinterwordspacing
T.~Linjordet and K.~Balog, ``Sanitizing synthetic training data generation for question answering over knowledge graphs,'' in \emph{Proceedings of the 2020 ACM SIGIR on International Conference on Theory of Information Retrieval}, ser. ICTIR '20.\hskip 1em plus 0.5em minus 0.4em\relax New York, NY, USA: Association for Computing Machinery, 2020, p. 121–128. [Online]. Available: \url{https://doi.org/10.1145/3409256.3409836}
\BIBentrySTDinterwordspacing

\bibitem{qald}
R.~Usbeck, R.~Gusmita, M.~Saleem, and A.-C. Ngonga~Ngomo, ``9th challenge on question answering over linked data (qald-9),'' 11 2018.

\bibitem{2019-SubQG}
J.~Ding, W.~Hu, Q.~Xu, and Y.~Qu, ``Leveraging frequent query substructures to generate formal queries for complex question answering,'' 01 2019, pp. 2614--2622.

\bibitem{modern_baselines}
\BIBentryALTinterwordspacing
D.~Banerjee, P.~A. Nair, J.~N. Kaur, R.~Usbeck, and C.~Biemann, ``Modern baselines for sparql semantic parsing,'' in \emph{Proceedings of the 45th International ACM SIGIR Conference on Research and Development in Information Retrieval}, ser. SIGIR '22.\hskip 1em plus 0.5em minus 0.4em\relax New York, NY, USA: Association for Computing Machinery, 2022, p. 2260–2265. [Online]. Available: \url{https://doi.org/10.1145/3477495.3531841}
\BIBentrySTDinterwordspacing

\bibitem{W3Crec-RDB-Direct-Mapping}
M.~Arenas, A.~Bertails, E.~{Prud'hommeaux}, and J.~Sequeda, ``A direct mapping of relational data to {RDF},'' {W3C} {Recommendation}, Sep. 2012, available at \protect\url{http://www.w3.org/TR/rdb-direct-mapping/}.

\bibitem{W3Crec-R2RML}
S.~Das, S.~Sundara, and R.~Cyganiak, ``{R2RML}: {RDB} to {RDF} mapping language,'' {W3C} {Recommendation}, Sep. 2012, available at \protect\url{http://www.w3.org/TR/r2rml/}.

\bibitem{CCKK*17}
D.~Calvanese, B.~Cogrel, S.~Komla-Ebri, R.~Kontchakov, D.~Lanti, M.~Rezk, M.~Rodriguez-Muro, and G.~Xiao, ``{Ontop}: {Answering} {SPARQL} queries over relational databases,'' vol.~8, no.~3, pp. 471--487, 2017.

\bibitem{IRNet}
\BIBentryALTinterwordspacing
J.~Guo, Z.~Zhan, Y.~Gao, Y.~Xiao, J.~Lou, T.~Liu, and D.~Zhang, ``Towards complex text-to-sql in cross-domain database with intermediate representation,'' \emph{CoRR}, vol. abs/1905.08205, 2019. [Online]. Available: \url{http://arxiv.org/abs/1905.08205}
\BIBentrySTDinterwordspacing

\bibitem{Text-to-text-transformer}
C.~Raffel, N.~Shazeer, A.~Roberts, K.~Lee, S.~Narang, M.~Matena, Y.~Zhou, W.~Li, and P.~J. Liu, ``Exploring the limits of transfer learning with a unified text-to-text transformer,'' \emph{J. Mach. Learn. Res.}, vol.~21, no.~1, jan 2020.

\bibitem{nadig-etal-2020-database}
\BIBentryALTinterwordspacing
S.~Nadig, M.~Braschler, and K.~Stockinger, ``\BIBforeignlanguage{English}{Database search vs. information retrieval: A novel method for studying natural language querying of semi-structured data},'' in \emph{\BIBforeignlanguage{English}{Proceedings of the Twelfth Language Resources and Evaluation Conference}}.\hskip 1em plus 0.5em minus 0.4em\relax Marseille, France: European Language Resources Association, May 2020, pp. 1772--1779. [Online]. Available: \url{https://aclanthology.org/2020.lrec-1.219}
\BIBentrySTDinterwordspacing

\end{thebibliography}

\end{document}